\pdfoutput=1

\documentclass[11pt]{article}

\usepackage{authblk}
\usepackage{acl2023}
\usepackage{svg}
\usepackage{times}
\usepackage{latexsym}
\usepackage{amsmath}
\usepackage{amssymb}
\usepackage{graphicx}
\usepackage{array}
\usepackage{multirow}
\usepackage{caption}
\usepackage{subcaption}

\usepackage[T1]{fontenc}

\usepackage[utf8]{inputenc}

\usepackage{microtype}

\usepackage{inconsolata}

\title{Domain Aligned Prefix Averaging for Domain Generalization in Abstractive Summarization}

\author[1]{\textbf{Pranav Ajit Nair}}
\author[1]{\textbf{Sukomal Pal}}
\author[2]{\textbf{Pradeepika Verma}}
\affil[1]{Indian Institute of Technology (BHU), Varanasi, India}
\affil[2]{TIH, Indian Institute of Technology, Patna, India}
\affil[1]{\texttt{\{pranavajitnair.cse18, spal.cse\}@itbhu.ac.in}}
\affil[2]{\texttt{pradeepikav.verma093@gmail.com}}

\begin{document}
\maketitle
\begin{abstract}
Domain generalization is hitherto an underexplored area applied in abstractive summarization. Moreover, most existing works on domain generalization have sophisticated training algorithms. In this paper, we propose a lightweight, weight averaging based, \textbf{D}omain \textbf{A}ligned \textbf{P}refix \textbf{A}veraging approach to domain generalization for abstractive summarization. Given a number of source domains, our method first trains a prefix for each one of them. These source prefixes generate summaries for a small number of target domain documents. The similarity of the generated summaries to their corresponding documents is used for calculating weights required to average source prefixes. In DAPA, prefix tuning allows for lightweight finetuning, and weight averaging allows for the computationally efficient addition of new source domains. When evaluated on four diverse summarization domains, DAPA shows comparable or better performance against the baselines, demonstrating the effectiveness of its prefix averaging scheme\footnote{\url{https://github.com/pranavajitnair/DAPA}}. 
\end{abstract}

\section{Introduction}

Abstractive document summarization aims at filtering the most crucial information in a document to present a concise view of it \citep{DBLP:conf/conll/NallapatiZSGX16, See2017GetTT}. This document may take the form of a news article \citep{hermann2015teaching}, a scientific paper \citep{Yasunaga2019ScisummNetAL}, a dialogue \citep{Gliwa2019SAMSumCA}, or a social media post \citep{DBLP:journals/corr/abs-1811-00783}.  The advent of pretrained models \citep{DBLP:journals/jmlr/RaffelSRLNMZLL20, DBLP:conf/acl/LewisLGGMLSZ20} has significantly improved abstractive summarization on several of the aforementioned domains. However, these approaches require extensive manual labelling of data which limits their use to domains without any labelled data. Given that real-world applications of summarization often face the problem of adapting to new domains, it becomes crucial to develop summarization systems that do well in data-scarce settings by leveraging information from the source domains.

Domain generalization accounts for learning a robust model for unseen domains from a set of source domains. This problem is closely related to transfer learning, multitask learning and domain adaptation all of which involve learning a model from a set of source tasks/domains to perform well on a set of target tasks/domains. However, in the case of domain generalization, labelled data for the target domain is unavailable. Previous works on domain generalization mainly focus on learning domain invariant features \citep{Gulrajani2020InSO, Wang2020UnseenTS, Li2018DomainGW}. Such methods work well on classification tasks where learning domain invariant features is sufficient to predict target classes. However, they may be insufficient for language generation tasks which have writing style, grammar as their ingredients \citep{DBLP:conf/acl/VuKPH22}. Moreover, such methods involve sophisticated algorithms for training and cannot be used for lightweight domain generalization.

Prefix tuning \citep{DBLP:conf/acl/LiL20} is a lightweight approach to adapting pretrained language models to downstream tasks. It augments the transformer self attention via prefix tokens learned through backpropagation on the task data while keeping the pretrained model's parameters frozen. Prompt tuning based approaches have been shown to do well on lifelong learning \citep{DBLP:journals/corr/abs-2110-07298} and zero-shot domain adaptation \citep{DBLP:conf/naacl/ZhaoZZHXJWW22}, which inspires us to adapt it to domain generalization for abstractive summarization. Concurrently, weight averaging has performed well on domain generalization tasks in Computer Vision \citep{DBLP:conf/nips/ChaCLCPLP21, Ram2022DiverseWA, DBLP:journals/corr/abs-2110-10832}. To improve functional diversity, these methods average model parameters from different runs and/or checkpoints.

\citet{DBLP:journals/corr/abs-2111-09832} applied weight averaging to NLP tasks. They merged models trained on different tasks/domains through fisher weight averaging. Their promising results motivate us to apply model merging through weight averaging for domain generalization. Keeping in mind the goal of generating a lightweight and parameter-efficient approach, and the benefits brought by weight averaging, we propose a lightweight \textbf{D}omain \textbf{A}ligned \textbf{P}refix \textbf{A}veraging, DAPA, approach to domain generalization for abstractive summarization. Our algorithm consists of three stages. First, prefixes are trained for each source domain. In the second stage, these source prefixes generate summaries for a small number of unlabelled target domain documents. In the third stage, the target domain prefix is obtained through a weighted average of these source prefixes. A higher document-summary similarity score, calculated from the summaries generated in the second stage, would assign a greater weightage to the corresponding source prefix. Through our prefix averaging scheme, we can identify source prefixes essential to ensure good performance on the target domain. Our extensive experimentation on four domains demonstrates the benefits bought by DAPA.

DAPA comes with the following advantages: i) It is a lightweight approach to domain generalization since the backbone pretrained model is frozen and only source prefixes are trained. ii) Through our novel prefix averaging scheme, DAPA is able to generalize well onto target domains. Moreover, freezing the backbone model's parameters further preserves generalization. iii) Our approach supports the efficient addition of new source domains since it only involves recomputing the prefix averaging weights.

To this end, we summarize our contributions as follows:
\begin{itemize}
    \item To the best of our knowledge, we are the first to explore prefix averaging for domain generalization on a language generation task.
    \item We propose a lightweight \textbf{D}omain \textbf{A}ligned \textbf{P}refix \textbf{A}veraging, DAPA, approach to domain generalization for abstractive summarization. DAPA first trains prefixes for each source domain, following which it utilizes the summary generation capabilities of these source prefixes to generalize to the target domain. 
    \item Through our experimentation setup we demonstrate the effectiveness of DAPA on domain generalization for abstractive summarization.
\end{itemize}

The rest of the paper is structured as follows: We explore related works in Section \ref{Related Work}. Section \ref{Method} describes our proposed approach DAPA. Section \ref{Experiments} and \ref{analysis} provide results for our domain generation experiments and a set of analysis we conduct on our approach. Section \ref{Conclusion} provides the conclusion and thoughts for future work. Section \ref{Limitations} discusses limitations and Section \ref{Ethics} sheds light om ethical risks of our work.

\section{Related Work} \label{Related Work}
\subsection{Abstractive Summarization}
Abstractive document summarization aims to distill the most critical information in a document to present a concise view of it. \citet{DBLP:conf/conll/NallapatiZSGX16} used an RNN based sequence to sequence model for abstractive summarization. \citet{See2017GetTT} used pointer generator networks to copy words from the input document. \citet{duan2019contrastive} augmented the transformer architecture with a contrastive attention mechanism to ignore the irrelevant parts of the document. \citet{zhang2020pegasus} pretrained a transformer model for summarization. \citet{DBLP:conf/acl/Liu020a} used contrastive loss for better re-ranking of summaries generated by pretrained models. \citet{paulus2018a} proposed the use of policy gradient reinforcement learning to alleviate exposure bias. \citet{gehrmann-etal-2018-bottom} developed a bottom-up copy attention mechanism to over-determine phrases in the document that should be included in the summary. Although great progress has been made in advancing state-of-the-art, few works have explored domain generalization for abstractive summarization. In this work, we develop a lightweight prefix averaging based method for domain generalization in abstractive summarization.

\begin{figure*}
     \centering
     \resizebox{\textwidth}{!}{%
         \includegraphics[width=\linewidth]{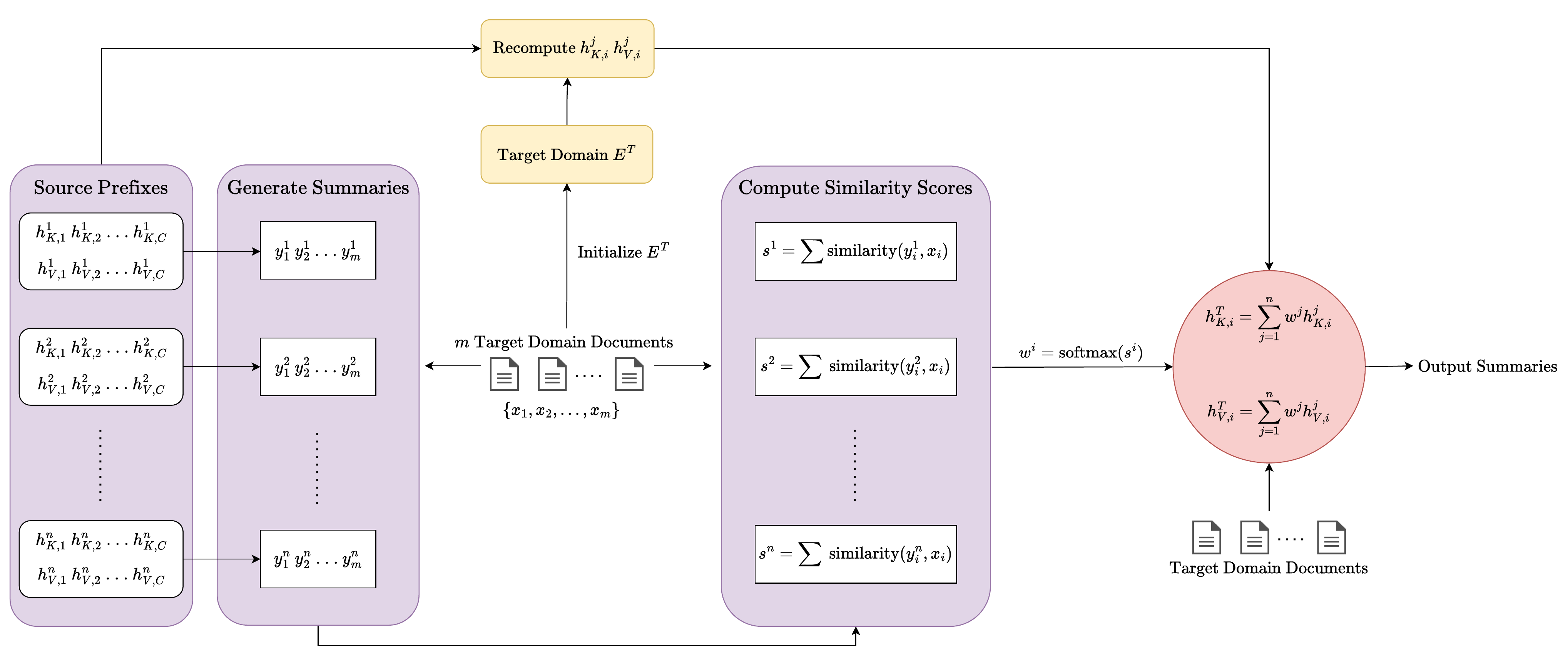}
         }
         \caption{Overview of the Domain Aligned Prefix Averaging model. Source prefixes are used for generating summaries for $m$ target domain documents. The similarity of these summaries to their corresponding documents are used for computing weights required to average source prefixes.}
        \label{fig:Fig0}
        
\end{figure*}

\subsection{Prompt Learning in Language Generation}
Prompts are task specific instructions prepended to the pretrained model's input. These task specific instructions are trained on the downstream task data while keeping the pretrained model's parameters frozen. \citet{DBLP:conf/acl/LiL20} proposed deep continuous prefixes that are prepended to the self attention layers of transformers. They demonstrated the effectiveness of using prefixes for language generation tasks such as summarization. \citet{DBLP:journals/corr/abs-2110-07298} prepended prompts into model embeddings for lifelong learning on language generation tasks. Similar to our approach, they train a separate prompt for each domain, however, their work does not focus on domain generalization. \citet{DBLP:journals/corr/abs-2110-06609} developed a multistage prompting network for machine translation where the encoder is prompted twice to refine the input representation, and the decoder is prompted once to generate the translation. \citet{DBLP:journals/corr/abs-2012-11926} used manually crafted templates for fixed-prompt tuning of pretrained models for few-shot summarization. \citet{DBLP:journals/corr/abs-2110-12680, DBLP:conf/naacl/DouLHJN21} used learnable prompts as guiding instructions for summarization. \citet{DBLP:conf/naacl/ZhaoZZHXJWW22}, similar to our approach, used prefixes to adapt to target domain. However, their approach involves pretraining prefix weights and cannot easily incorporate new source domains. On the other hand, our work takes a weighted average of source prefixes and can easily add new source domains to the mix.

\subsection{Weight Averaging and Domain Generalization}
Domain generalization accounts for learning a robust model for unseen domains from a set of source domains. It has mostly been studied in Computer Vision.  The main approaches are based on invariant feature learning \citep{Li2018DomainGW, Wang2020UnseenTS}, data augmentation \citep{Wang2020MetaLearningFD}, and meta learning \citep{DBLP:conf/icassp/WangLK20}. For language generation, \citet{DBLP:conf/acl/VuKPH22} proposed a leave-one-domain-out strategy to fuse adaptors for machine translation. Recently, weight averaging \citep{DBLP:conf/nips/GaripovIPVW18} has been successfully applied to domain generalization. \citet{DBLP:conf/nips/ChaCLCPLP21} utilized weight averaging to attain a flat minima, and thereby attained generalization across domains. \citet{Ram2022DiverseWA} averaged weights across multiple runs, trained with different hyperparameters to improve domain generalization. \citet{DBLP:journals/corr/abs-2110-10832} ensembled a moving average of weights across multiple checkpoints. Unlike these approaches, we develop a novel mechanism to generate weights for averaging source prefixes, and evaluate our approach on abstractive summarization. \citet{DBLP:journals/corr/abs-2111-09832} also developed a scheme to average model weights. They utilized Fisher information in model parameters for weight averaging. However, they only evaluated on NLU tasks.

\section{Method} \label{Method}
We first describe the domain generalization problem in Section \ref{PD}. Then we move on to describe prefix tuning in Section \ref{PT}. Finally, in Section \ref{DAPA main}, we describe our proposed approach, DAPA.

\subsection{Problem Definition} \label{PD}
Let $D^S = \{D_{1}^{S}, D_{2}^{S}, . . . , D_{n}^{S}\}$ be the set of source domains. We denote the target domain with $D^T$. Domain generalization aims to seek a network which generalizes well on $D^T$ when trained on $D^S$. We require our model to generate fluent summaries for target domain documents when trained on source domain documents.

\subsection{Prefix Tuning} \label{PT}
We utilize prefix tuning to train a separate prefix for each source domain. We begin by restating the transformer attention:
\begin{equation}
     \text{attn}(Q, K, V) = \text{softmax} (\frac{Q\cdot K^{\top}}{\sqrt{d}})V
\end{equation}
Here, the query matrix $Q$, the key matrix $K$, and the value matrix $V$ are obtained through independent linear transformations on the output of the previous layer/encoder. $d$ is the model dimension. Note that we omit the multihead notation for clarity.

Prefix tuning modifies the transformer attention by adding tunable prefixes to $K$ and $V$. Consequently $K$ is modified as $K^{\prime} = [h_K;K]$ and $V$ is modified as $V^{\prime} = [h_V;V]$. Here $h_K$ and $h_V$ represent the key prefix and the value prefix respectively.

Following \citet{DBLP:conf/acl/LiL20}, we model these prefixes using a two layer MLP as follows:
\begin{equation}
\label{prefix eq}
    \begin{aligned}
       h^{j}_K=W_{K,2}^jf(W_{K,1}^j E^j+b_{K,1}^j)+b_{K,2}^j \\
       h^{j}_V=W_{V,2}^jf(W_{V,1}^j E^j+b_{V,1}^j)+b_{V,2}^j
    \end{aligned}
\end{equation}
where $W_{K,1}^j, W_{V,1}^j, W_{K,2}^j, W_{V,2}^j \in  \mathbb{R}^{d\times d}$  are trainable weights, and $b_{K,1}^j, b_{V,1}^j, b_{K,2}^j, b_{V,2}^j \in  \mathbb{R}^{d}$ are trainable biases. $E^j \in \mathbb{R}^{C \times d}$ is a trainable embedding matrix with $C$ as the prefix length. Index $j$ corresponds to source domain $D_{j}^{S}$. We detail the initialization of $E^j$ in Section \ref{training details}. Each source prefix is trained in an end-to-end fashion on its corresponding source domain data.

\subsection{Domain Aligned Prefix Averaging}
\label{DAPA main}
Having formulated our problem and described prefix tuning, we now describe our approach, DAPA.

\subsubsection{Computing Weights to Average Source Prefixes} \label{DAPA}
DAPA utilizes the summary generation capabilities of source prefixes to generate weights for averaging source prefixes. Let $D^{T}_{m, sample} = \{x_1, x_2, . . . , x_m\}$ be a set $m$ unlabelled documents from the target domain. In our experiments, we observe that a value of $m$ as small as $50$ suffices. Let $P^S = \{P_{1}^{S}, P_{2}^{S}, . . . , P_{n}^{S}\}$ represent the set of source prefixes. Note that $P^{S}_{j} = \{ h_{K}^j, h_{V}^j \}$. For target domain document $x_i$, DAPA first generates $n$ summaries pertaining to each source prefix as follows:
\begin{equation}
    y_{i}^{j}  = M(x_i; P_{j}^{S}) \text{;  }  \text{ } 1 \leq i \leq m \text{,  }  \text{ } 1 \leq j \leq n
\end{equation}
where $M$ represents the frozen pretrained language model. 

Next, it uses an encoder $f$ to generate sentence representations for the summary $y_{i}^{j}$ as $r_{i}^{j} = f(y_{i}^{j})$ and the document $x_{i}$ as $t_{i} = f(x_{i})$. We use SentenceBERT \citep{Reimers2019SentenceBERTSE} as our encoder. Following this, we compute average document-summary cosine similarity scores for each source prefix as follows:
\begin{equation}
\label{eq 4}
    s^j = \sum_{i=1}^{m} \text{cosine-similarity}(r_{i}^{j}, t_{i})
\end{equation}
The final weights for averaging source prefixes are generated by taking a softmax over the average document-summary similarity scores as:
\begin{equation}
\label{eq 5}
    w^j = \frac{\text{exp}(s^j)}{\sum_{i = 1}^{n} \text{exp}(s^i)}
\end{equation}

\subsubsection{Prefix Averaging} \label{TTA}
Given a target domain document $x$, we wish to generate a summary $y$ with a target prefix obtained by averaging the source prefixes using our weight averaging scheme described in Section \ref{DAPA}. Through $W = \{w^1, w^2, . . . , w^n\}$, we take a weighted average of the source prefixes as follows:
\begin{equation}
    \begin{aligned}
        h^{T}_{K} = \sum_{j = 1}^{n} w^j h_{K}^j \\
        h^{T}_{V} = \sum_{j = 1}^{n} w^j h_{V}^j \\
        P^T = \{ h_{K}^T, h_{V}^T \}
    \end{aligned}
\end{equation}
where $P^T$ is the target prefix through which the target summary $y = M(x; P^T)$ is generated. Note that test time averaging requires recomputation of $h_{K}^j$ and $h_{V}^j$ by replacing $E^j$ with $E^T$ in equation \ref{prefix eq}. We detail the computation of $E^T$ in Section \ref{training details}.

\section{Experimental Setup} \label{Experiments}

\subsection{Dataset and Metrics}
We use four summarization datasets, each belonging to a different domain. For the \textit{news} domain we use the CNN/Daily Mail dataset \citep{hermann2015teaching}; Samsum \citep{Gliwa2019SAMSumCA} for the \textit{chat} domain; Reddit posts for the \textit{social-media} domain \citep{DBLP:journals/corr/abs-1811-00783}; ScisummNet \citep{Yasunaga2019ScisummNetAL} for training on the \textit{scientific} domain and Cl-SciSumm \citep{DBLP:conf/sigir/JaidkaYCRK18} for testing on the \textit{scientific} domain. Dataset statistics are presented in Table \ref{tab:Table 1}. For evaluation, we report ROUGE-1, ROUGE-2 and ROUGE-L metrics\footnote{\url{https://github.com/pltrdy/files2rouge}} \citep{Lin2004ROUGEAP}.
\begin{table}
\resizebox{\columnwidth}{!}{%
 \begin{tabular}{|l|l|l|l|} 
 \hline
  \textbf{Domain} & \textbf{Train size} & \textbf{Dev size} & \textbf{Test size} \\
 \hline
    \textit{News} & 287,113 & 13,368 & 11,490
 \\ 
     \textit{Scientific } & 947 &  10 &   10 
 \\
      \textit{Social-media} & 33246 & 4151 & 4155
 \\
      \textit{Chat} & 14,732 & 818 & 819
\\
 \hline
\end{tabular}
}
\caption{Dataset statistics for each domain.}

\label{tab:Table 1}
\end{table}

\begin{table*}[]
\centering
\resizebox{\textwidth}{!}{%
\begin{tabular}{|c|ccc|ccc|ccc|ccc|}
\hline
\multirow{2}{*}{Approach} & \multicolumn{3}{c|}{News} & \multicolumn{3}{c|}{Scientific} & \multicolumn{3}{c|}{Chat} & \multicolumn{3}{c|}{Social-media} \\ \cline{2-13} 

                          & R-1     & R-2     & R-L     & R-1       & R-2       & R-L       & R-1      & R-2    & R-L     & R-1        & R-2       & R-L        \\\hline
ERM-finetune              & 39.94  & 18.09  & \textbf{32.90}   & \textbf{33.72}    & \textbf{22.14}    & \textbf{31.66}    & 25.52   & 6.69  & 21.09  & 16.53     & 2.94     & 11.58     \\
ERM-prefix                & 38.05  & 16.49  & 32.15  & 29.29    & 17.55    & 26.78    & 22.46   & 5.47  & 19.39  & \textbf{17.5}      & \textbf{3.01}     & \textbf{12.36}     \\
DAPA-average              & 35.98  & 15.90   & 30.58  & 28.78    & 17.04    & 26.52    & 23.04   & 5.55  & 19.68  & 16.86     & 2.88     & 11.88     \\
DAPA-max                  & 35.18  & 15.39  & 30.36  & 28.57    & 15.84    & 25.03    & 26.64   & 7.32  & 22.19  & 16.97     & 2.76     & 11.83     \\
\textbf{DAPA-inst}                 & 35.93  & 15.91  & 30.58  & 29.34    & 18.72    & 27.40     & 20.81   & 4.78  & 17.79  & 16.54     & 2.81     & 11.64     \\
\textbf{DAPA }                     & \textbf{40.28}  & \textbf{18.12}  & 32.78  & 30.84    & 18.97    & 27.23    & \textbf{28.23}   & \textbf{8.70}   & \textbf{22.86 } & 14.48     & 2.68     & 9.70   \\
\hline
\end{tabular}
}
\caption{ROUGE scores for domain generalization on the four domains.}
\label{tab:Table 2}
\end{table*}

\begin{table*}
\begin{minipage}{.48\linewidth}
\centering
\resizebox{\columnwidth}{!}{%
\begin{tabular}{|c|c|c|c|c|}
\hline
Target                  & Excluded  & \multirow{2}{*}{R-1} & \multirow{2}{*}{R-2} & \multirow{2}{*}{R-L} \\
Domain & Domain & & & \\
\hline
\multirow{3}{*}{Scientific}   & News            & 30.61   & 18.01   & 26.46   \\
                              & Chat            & 30.20    & 19.54   & 26.78   \\
                              & Social-media    & 31.25   & 19.26   & 27.26   \\
\hline
\multirow{3}{*}{Chat}         & News            & 28.23   & 8.67    & 22.86   \\
                              & Social-media    & 28.23   & 8.67    & 22.86   \\
                              & Scientific      & 25.63   & 6.51    & 21.16   \\
\hline
\multirow{3}{*}{Social-media} & News            & 14.6    & 2.71    & 9.79    \\
                              & Chat            & 14.6    & 2.71    & 9.79    \\
                              & Scientific      & 17.25   & 3.04    & 12.24   \\
\hline
\multirow{3}{*}{News}         & Social-media    & 40.29   & 18.12   & 32.78   \\
                              & Chat            & 40.27   & 18.12   & 32.78   \\
                              & Scientific      & 38.06   & 16.71   & 32.28  \\
\hline
\end{tabular}
}
\caption{ROUGE scores while using only two source domains to compute $W$. The second column mentions the domain whose prefix has been left out for the target domain's prefix computation.}
\label{tab:Table 3}
\end{minipage}
\hfill
\begin{minipage}{.48\linewidth}
\centering
\resizebox{\columnwidth}{!}{%
\begin{tabular}{|c|c|c|c|c|}
\hline
Target                   & Source & \multirow{2}{*}{R-1} & \multirow{2}{*}{R-2} & \multirow{2}{*}{R-L} \\
Domain & Domain & & & \\
\hline
\multirow{3}{*}{Scientific}   & News          & 36.73   & 25.01   & 34.07   \\
                              & Chat          & 32.14   & 19.08   & 28.15   \\
                              & Social-media  & 18.81   & 9.42    & 17.40    \\
\hline
\multirow{3}{*}{Chat}         & News          & 25.61   & 6.49    & 21.11   \\
                              & Social-media  & 21.21   & 5.76    & 18.87   \\
                              & Scientific    & 28.23   & 8.70     & 22.86   \\
\hline
\multirow{3}{*}{Social-media}  & News          & 17.22   & 3.01    & 12.23   \\
                               & Chat          & 16.37   & 2.72    & 11.61   \\
                               & Scientific    & 14.61   & 2.71    & 9.79    \\
\hline
\multirow{3}{*}{News}         & Social-media  & 30.47   & 13.49   & 27.14 \\
                              & Chat          & 38.03   & 16.69   & 32.25  \\
                              & Scientific    & 40.27   & 18.12   & 32.78   \\
\hline
\end{tabular}
}
\caption{ROUGE scores while using a single source domain's prefix to generate target domain summaries. The second column mentions the source domain adopted to generate summaries on the target domain.}
\label{tab:Table 4}
\end{minipage}
\end{table*}
\subsection{Baselines}
We use the method of empirical risk minimization (ERM) as our primary baseline. It trains the model by minimizing the sum of errors across source domains and examples. For computer vision tasks, \citet{DBLP:conf/iclr/GulrajaniL21} have shown that a well tuned ERM baseline performs competitively with several sophisticated methods for domain generalization. Thus, we use it as our primary baseline. We define two variants of ERM: i) ERM-finetune, finetunes the pretrained language model on a combination of all source domains. ii) ERM-prefix, prefix-tunes the backbone language model on a combination of all source domains. To validate the efficacy of our weight averaging scheme, we create two variants of DAPA, namely DAPA-average and DAPA-max. DAPA-average follows $w^j = \frac{1}{n}$ and DAPA-max takes a max pooling operation over the source prefixes instead of averaging them. We also present results for an instantaneous version of DAPA, DAPA-inst. Here, instead of using $D^T_{m, sample}$, we use the current testing document to compute weights $W$. We also consider four additional baselines as an upper bound to our method, results for which are presented in Appendix \ref{Additional Results}.   

\begin{figure*}
     \centering
     \begin{subfigure}[b]{0.32\textwidth}
         \includegraphics[width=\textwidth]{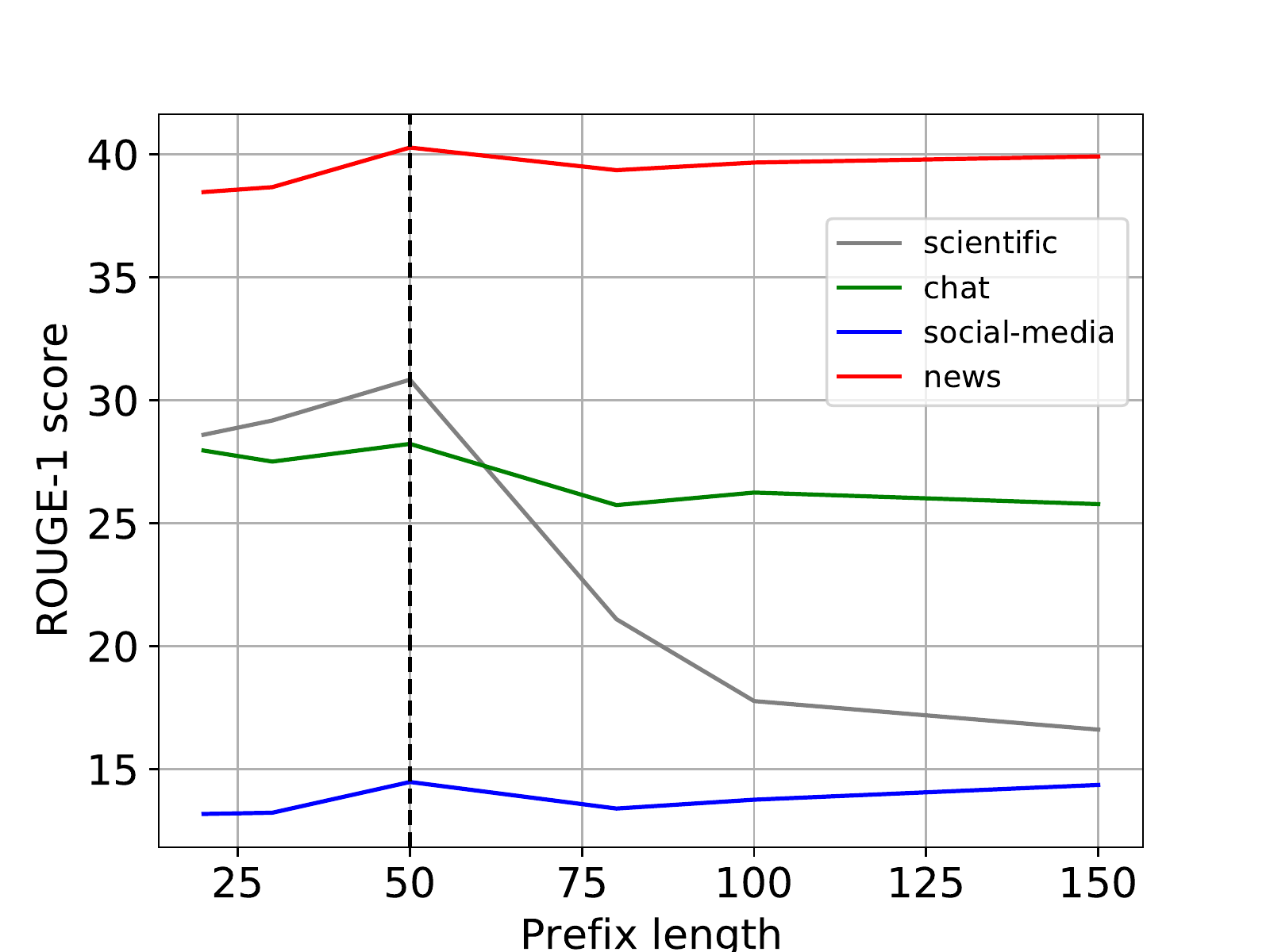}
         \caption{ROUGE-1 variation.}
     \end{subfigure}
     \begin{subfigure}[b]{0.32\textwidth}
         \includegraphics[width=\textwidth]{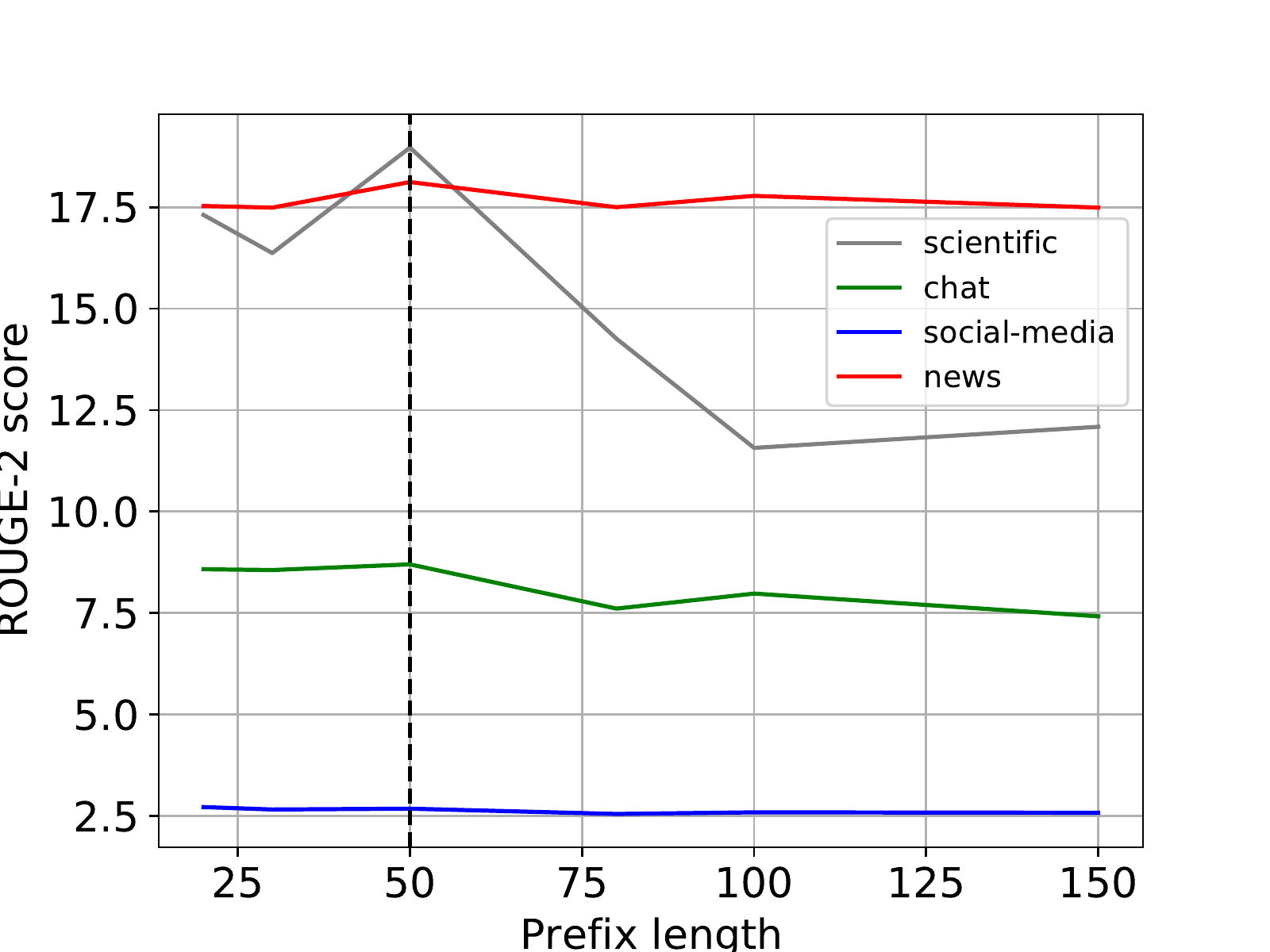}
         \caption{ROUGE-2 variation.}
     \end{subfigure}
     \begin{subfigure}[b]{0.32\textwidth}
         \includegraphics[width=\textwidth]{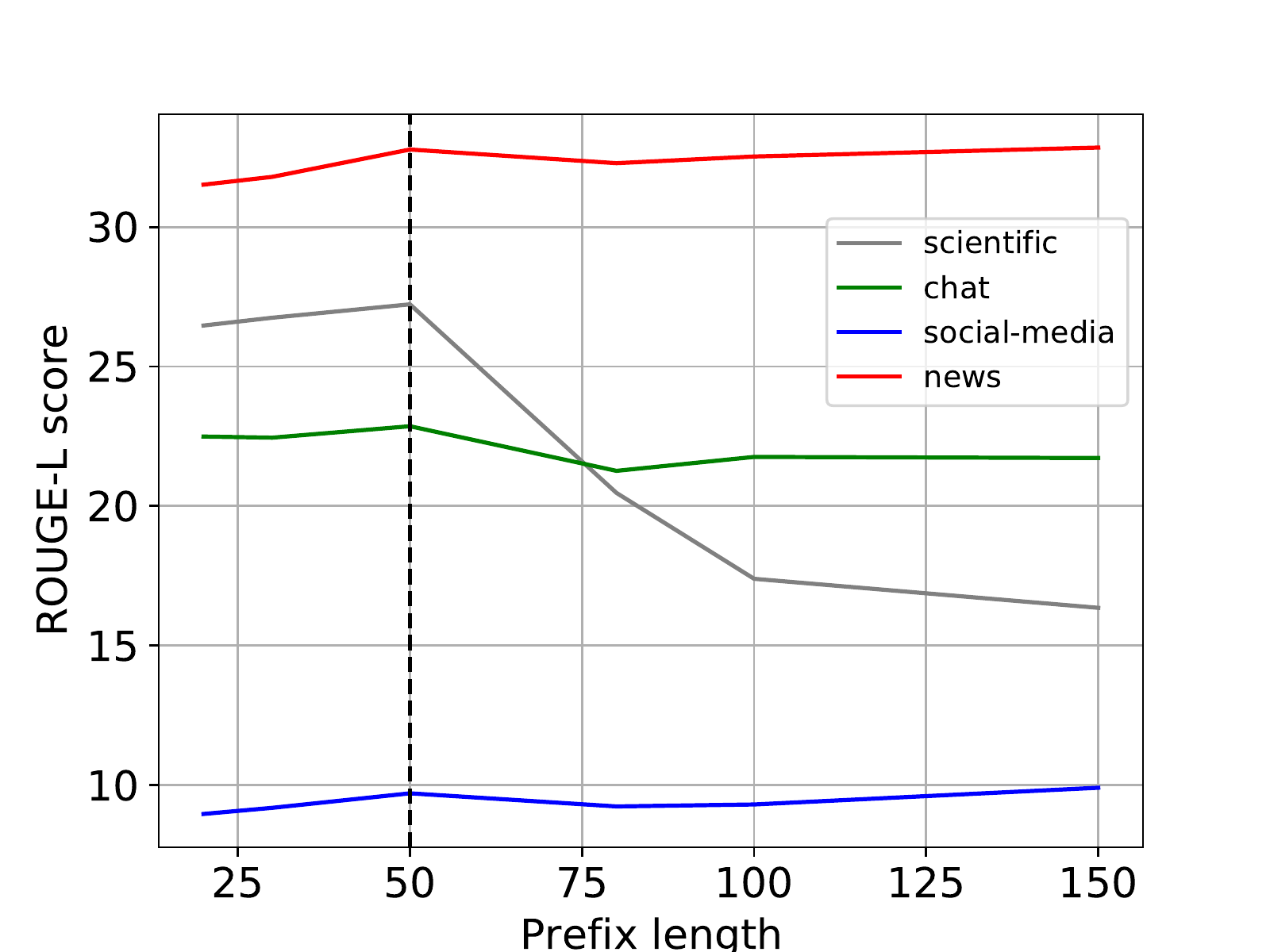}
         \caption{ROUGE-L variation.}
     \end{subfigure}
        \caption{Variation of ROUGE scores with prefix length $C$. Empirically, $C = 50$ is the most optimal prefix length. Throughout this experiment, $m = 50$.}
        \label{fig:Fig1}
\end{figure*}

\subsection{Training Details} \label{training details}
For our backbone pretrained model, we use T5-small (containing roughly 60M parameters) \citep{DBLP:journals/jmlr/RaffelSRLNMZLL20}. Prefix tuning adds rough 922K test time parameters to T5-small. Prefix length $C$ is fixed to $50$ unless otherwise specified. $m$ is also set to $50$. We verify our choices for $C$ and $m$ in Section \ref{analysis}. Both, finetuning and prefix tuning experiments are optimized with Adafactor \citep{DBLP:conf/icml/ShazeerS18}. Finetuning uses a maximum learning rate of $5e-4$, a square root decay schedule, and a linear warmup of $5000$ steps. Prefix tuning uses a constant learning rate of $5e-3$. All other Adafactor specific hyperparameters are left to their default values in HuggingFace-transformers\footnote{\url{https://github.com/huggingface/transformers}} \citep{wolf-etal-2020-transformers}. We utilize OpenPrompt\footnote{\url{https://github.com/thunlp/OpenPrompt}} \cite{DBLP:conf/acl/DingHZCLZS22} and HuggingFace-transformers to implement prefix tuning, and use the sentence-transformers\footnote{\url{https://huggingface.co/sentence-transformers/all-MiniLM-L6-v2}} implementation for SentenceBERT.

For finetuning, we employ a batch size of $5$ with gradient accumulation up to $5$ iterations. For prefix tuning, we use a batch size of $5$ but without any gradient accumulation. All our experiments are run on a single Nvidia-RTX 2080 Ti machine. One finetuning weight update (via gradient accumulation) takes rough 224 milliseconds and one prefix tuning iteration takes roughly 139 milliseconds. All our models are trained for 10 epochs with early stopping performed through validation ROUGE scores. For ERM-finetune and ERM-prefix, the training process is stopped if the in-domain validation scores for any of the three source domains starts to fall. Each training experiment is carried out only once.

\begin{figure*}
     \centering
     \begin{subfigure}[b]{0.32\textwidth}
         \includegraphics[width=\textwidth]{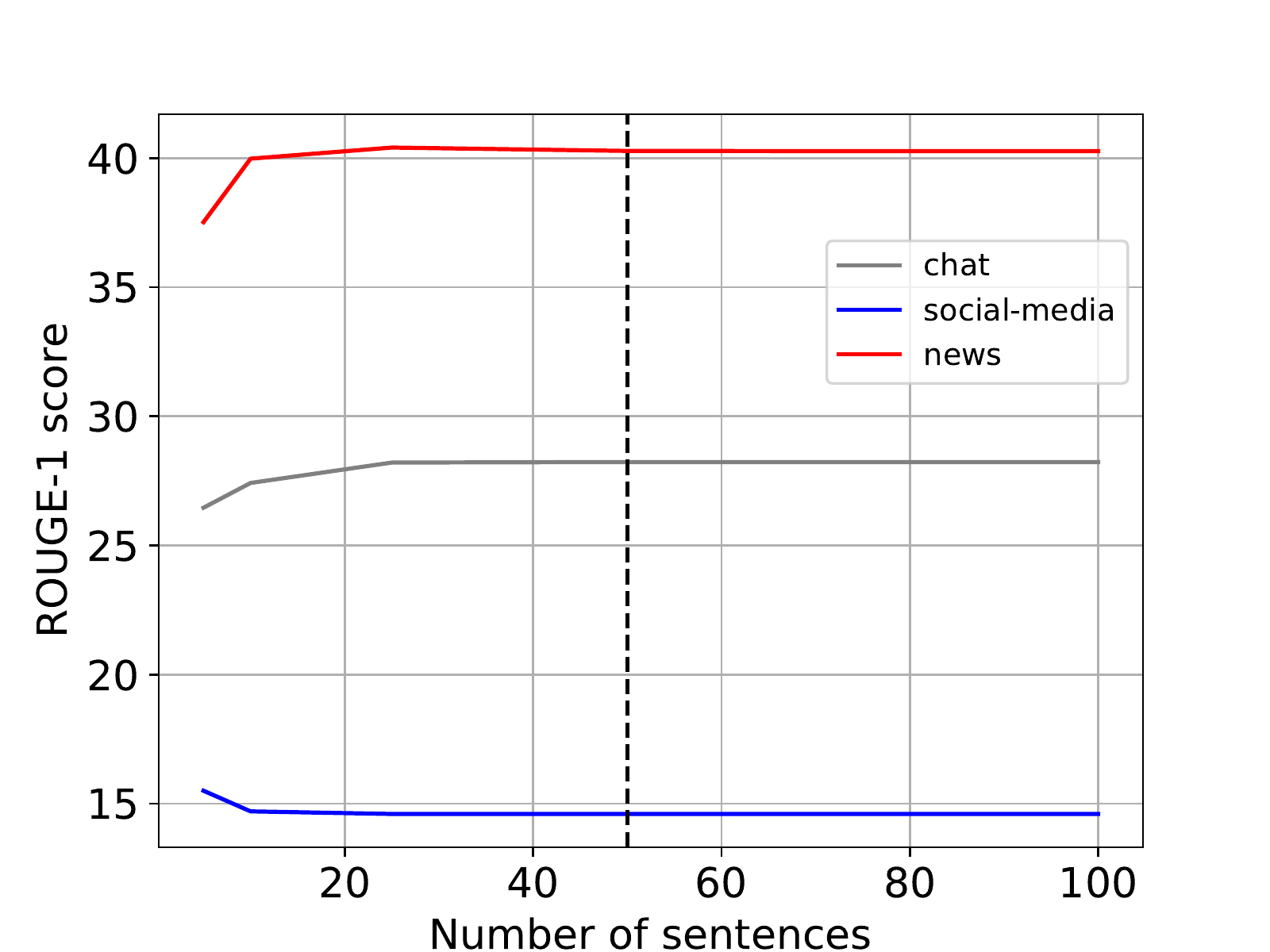}
         \caption{ROUGE-1 variation.}
     \end{subfigure}
     \begin{subfigure}[b]{0.32\textwidth}
         \includegraphics[width=\textwidth]{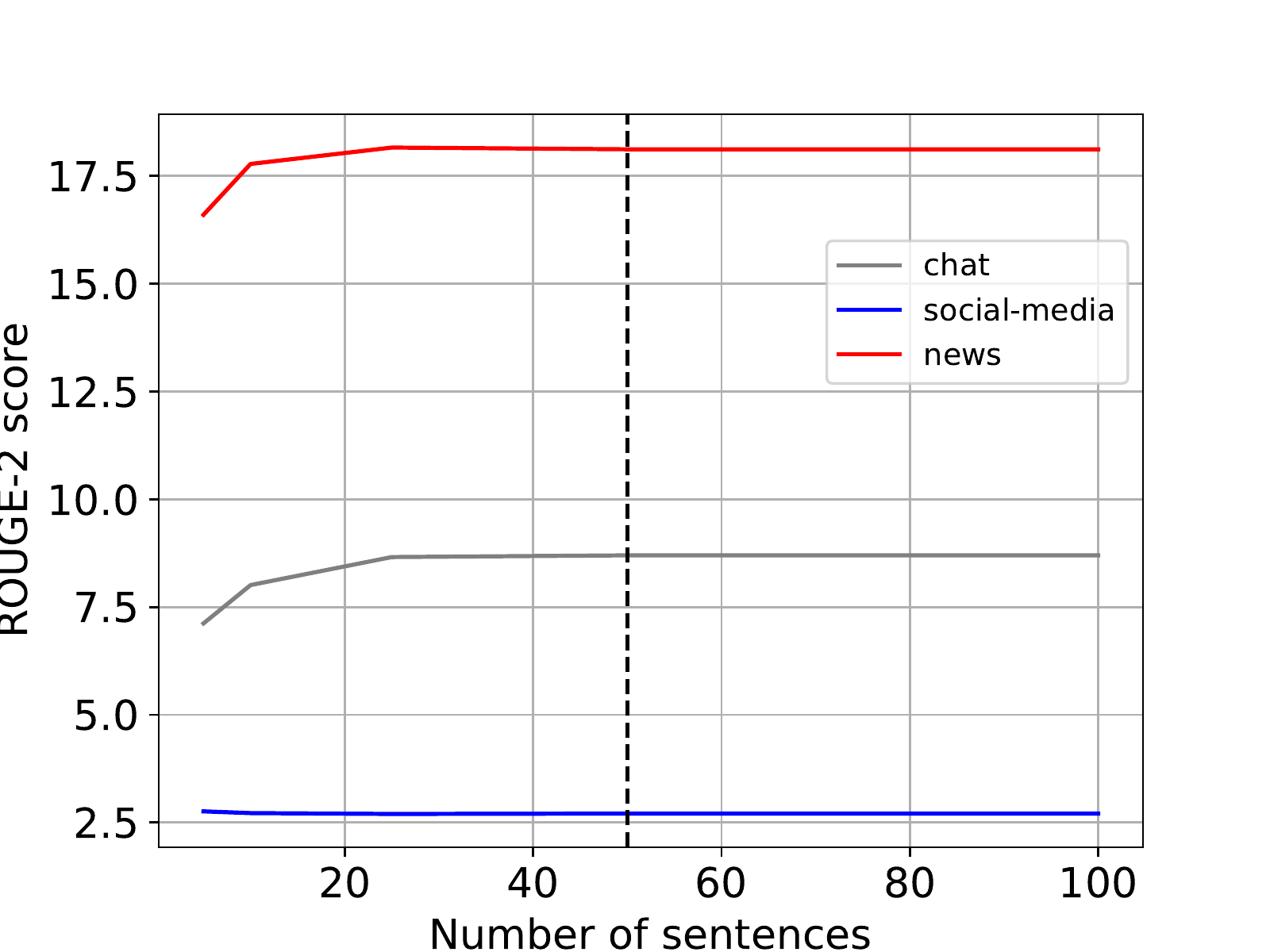}
         \caption{ROUGE-2 variation.}
     \end{subfigure}
     \begin{subfigure}[b]{0.32\textwidth}
         \includegraphics[width=\textwidth]{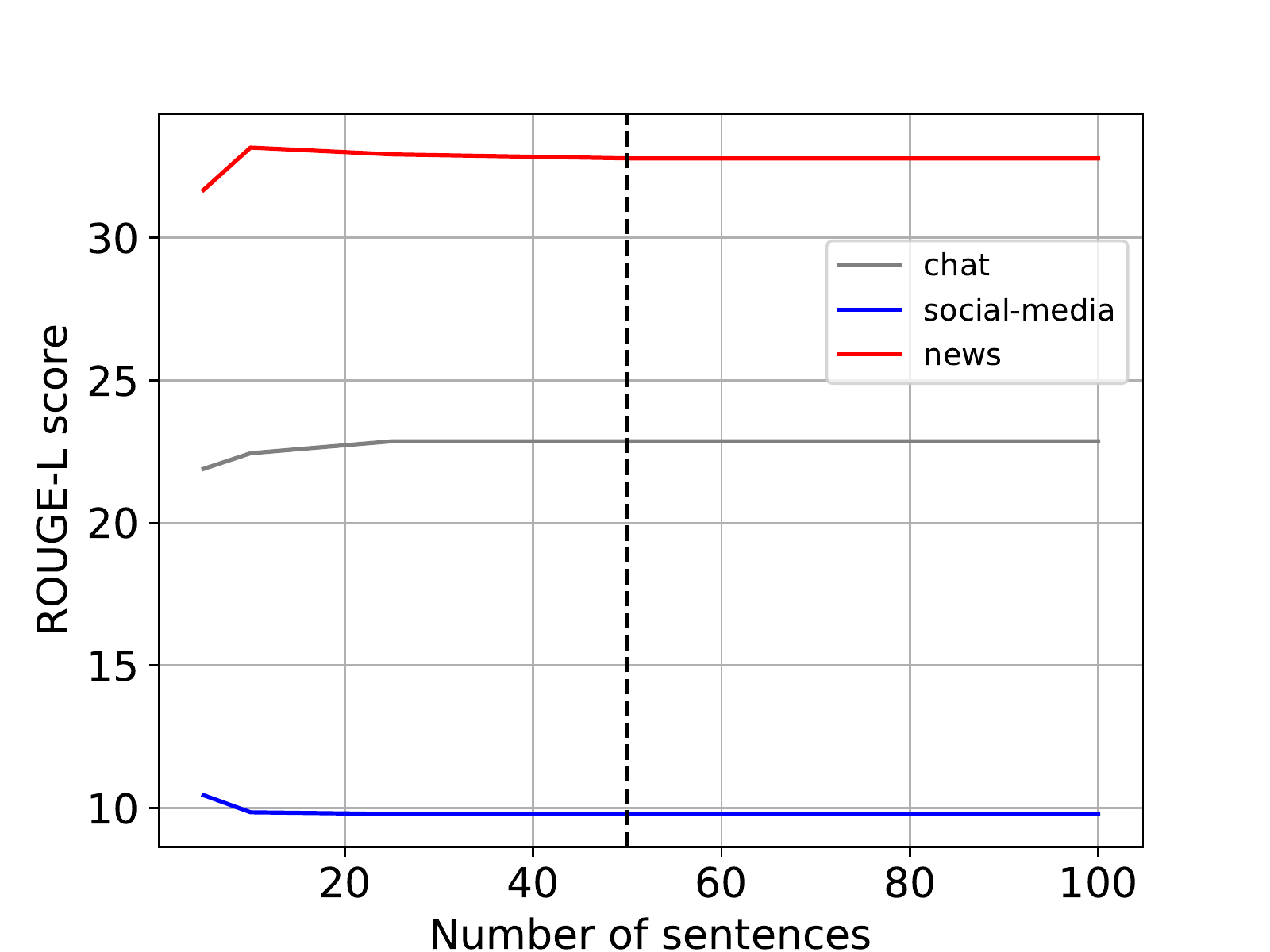}
         \caption{ROUGE-L variation.}
     \end{subfigure}
        \caption{Variation of ROUGE scores with the number of sentences used for computing weights $W$. Throughout this experiment, $C = 50$ and $E^T$ is initialized with  $50$ most frequent tokens from $D^T_{m, sample}$.}
        \label{fig:Fig 2}
\end{figure*}
\begin{figure*}
     \centering
     \begin{subfigure}[b]{0.32\textwidth}
         \includegraphics[width=\textwidth]{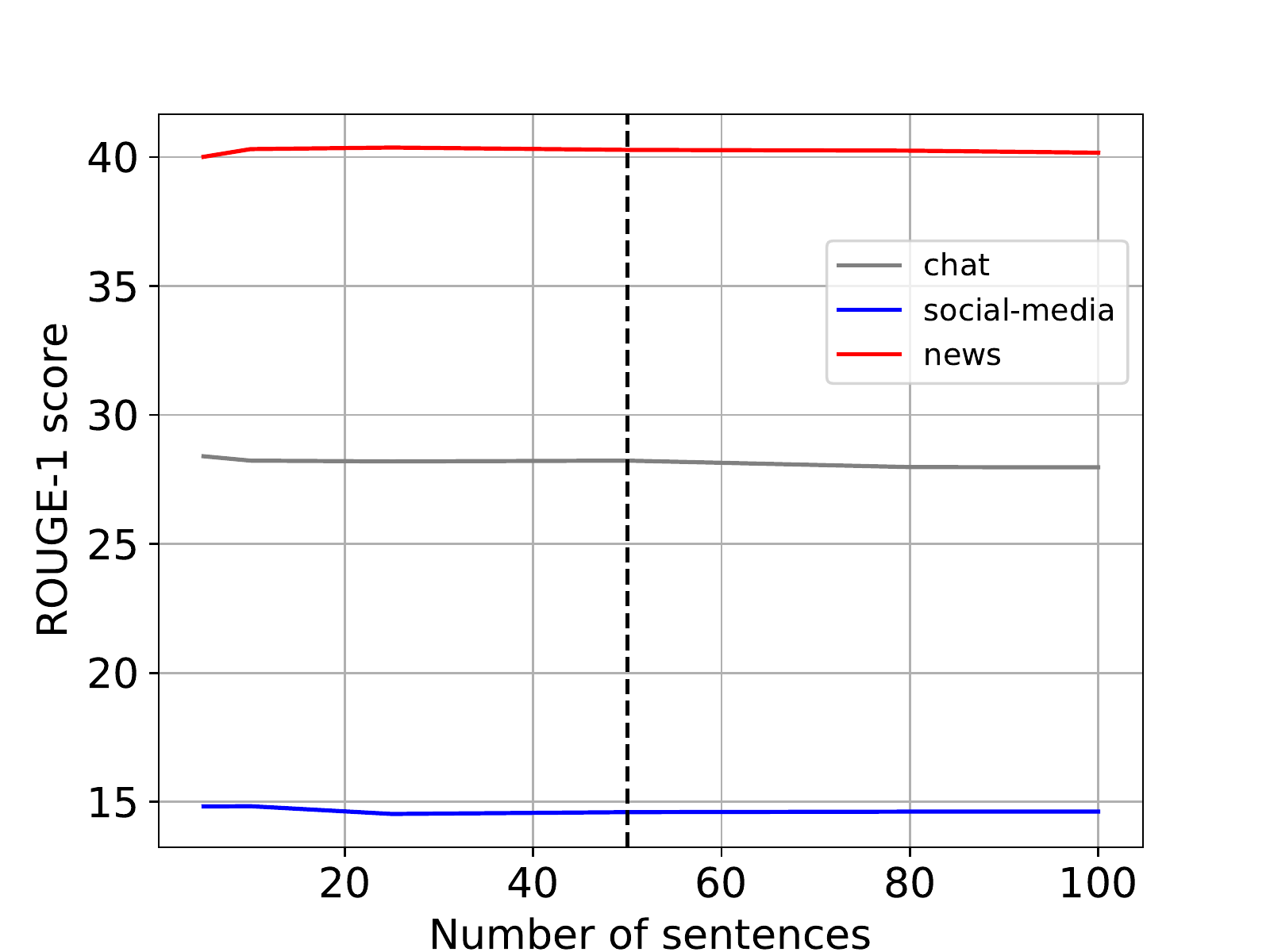}
         \caption{ROUGE-1 variation.}
     \end{subfigure}
     \begin{subfigure}[b]{0.32\textwidth}
         \includegraphics[width=\textwidth]{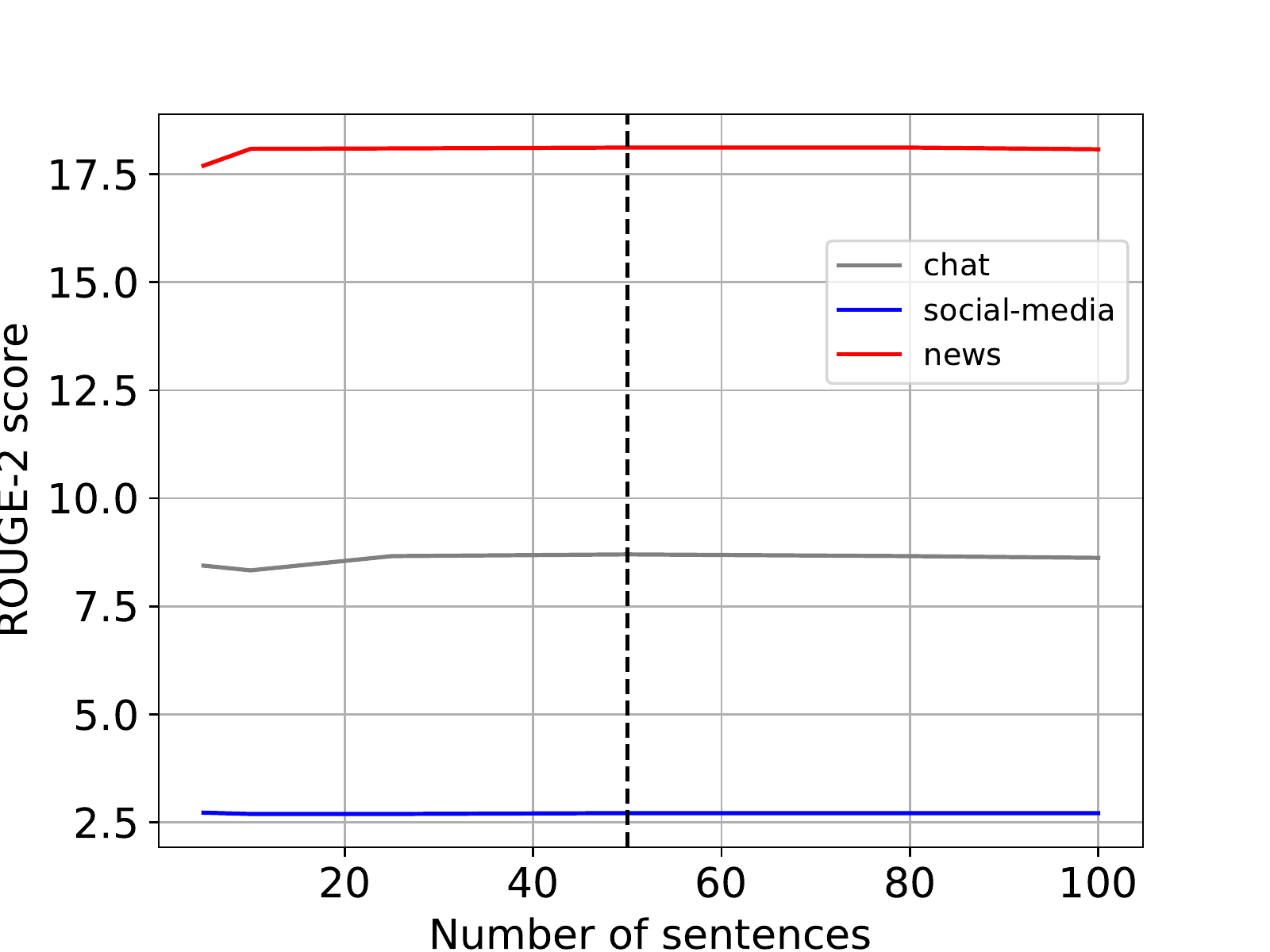}
         \caption{ROUGE-2 variation.}
     \end{subfigure}
     \begin{subfigure}[b]{0.32\textwidth}
         \includegraphics[width=\textwidth]{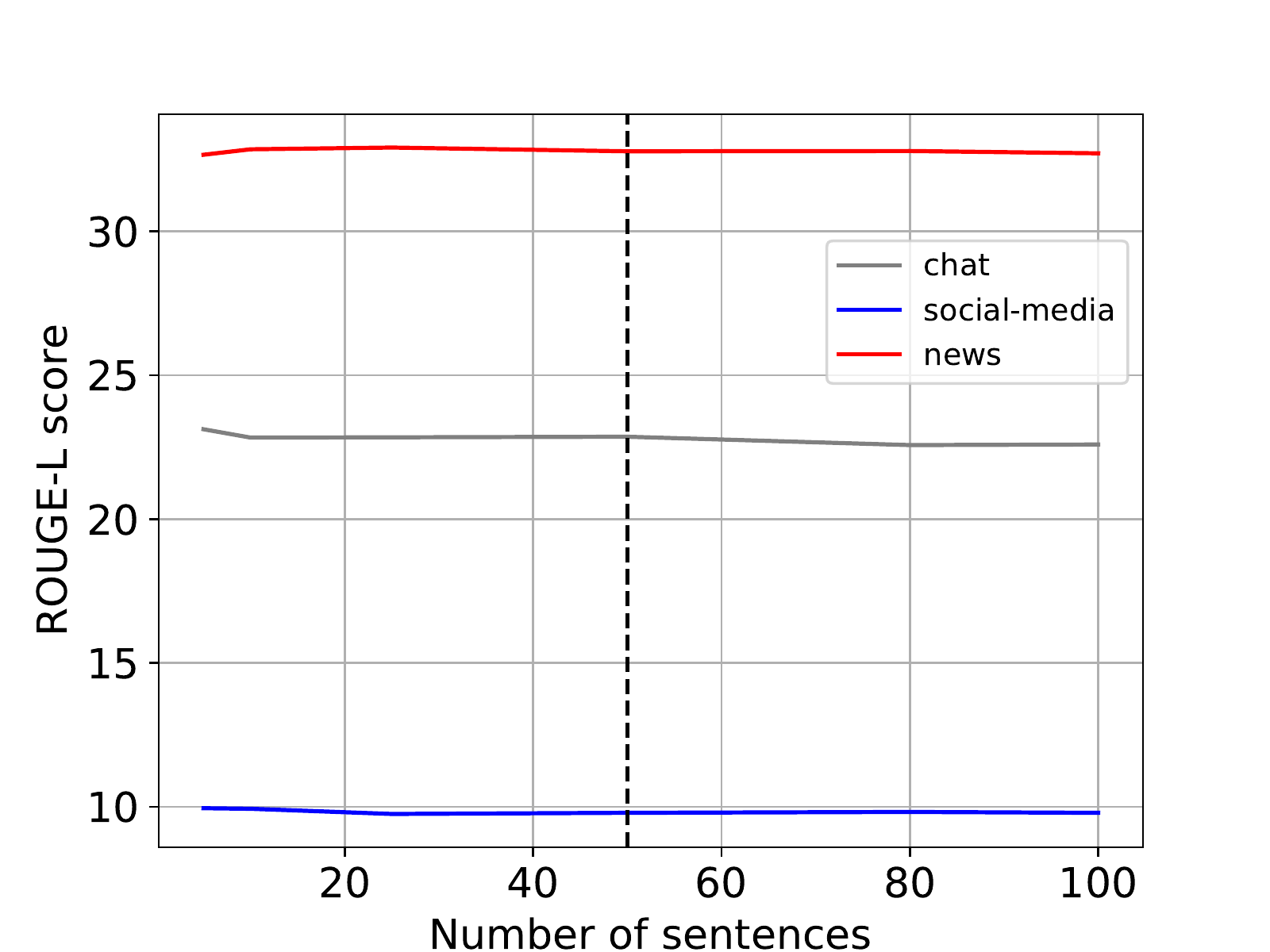}
         \caption{ROUGE-L variation.}
     \end{subfigure}
        \caption{Variation of ROUGE scores with the number of sentences used for deriving the $C$ most frequent words. Throughout this experiment, $C = 50$.}
        \label{fig:Fig3}
\end{figure*}

For prefix tuning, we initialize $E^T$ with T5 embeddings of the $C$ most frequent sentencepiece\footnote{\url{https://github.com/google/sentencepiece}} \citep{DBLP:conf/emnlp/KudoR18} tokens of $D^T_{m, sample}$. For DAPA-inst, the same process is applied, however instead of $D^T_{m, sample}$, the current test document is used. For source domains, $C$ most frequent tokens are extracted from the train set.

Summary generation uses a beam length of $10$ and a repetition penalty of $2.5$. All source documents are truncated to $512$ sentencepiece tokens and all summaries are truncated to $200$ sentencepiece tokens. For the \textit{scientific} domain, we only include the document's abstract, introduction and conclusion in our input to present the document's most crucial aspects within T5's maximum allowed sequence length. All our results are presented on the test sets of the four domains.

\subsection{Main Results}
\label{main results}
Table \ref{tab:Table 2} presents results for our domain generalization experiments. DAPA outperforms all compared methods on the \textit{chat} domain and outperforms all compared methods on two out of the three ROUGE scores on the \textit{news} domain. Owing to our prefix averaging method, DAPA demonstrates better generalization capabilities when compared to ERM-finetune and ERM-prefix. DAPA-max and DAPA-average do not utilize the summary generation capabilities of source prefixes and, thus, fail to account for the aspects most crucial to summary generation for the two domains. Also, DAPA-inst performs significantly worse than DAPA, thereby emphasizing the importance of using a greater number of target domain documents to better approximate the weights for averaging source prefixes.

\begin{figure*}
     \begin{subfigure}[t]{.3\linewidth}
         \includegraphics[width=\textwidth]{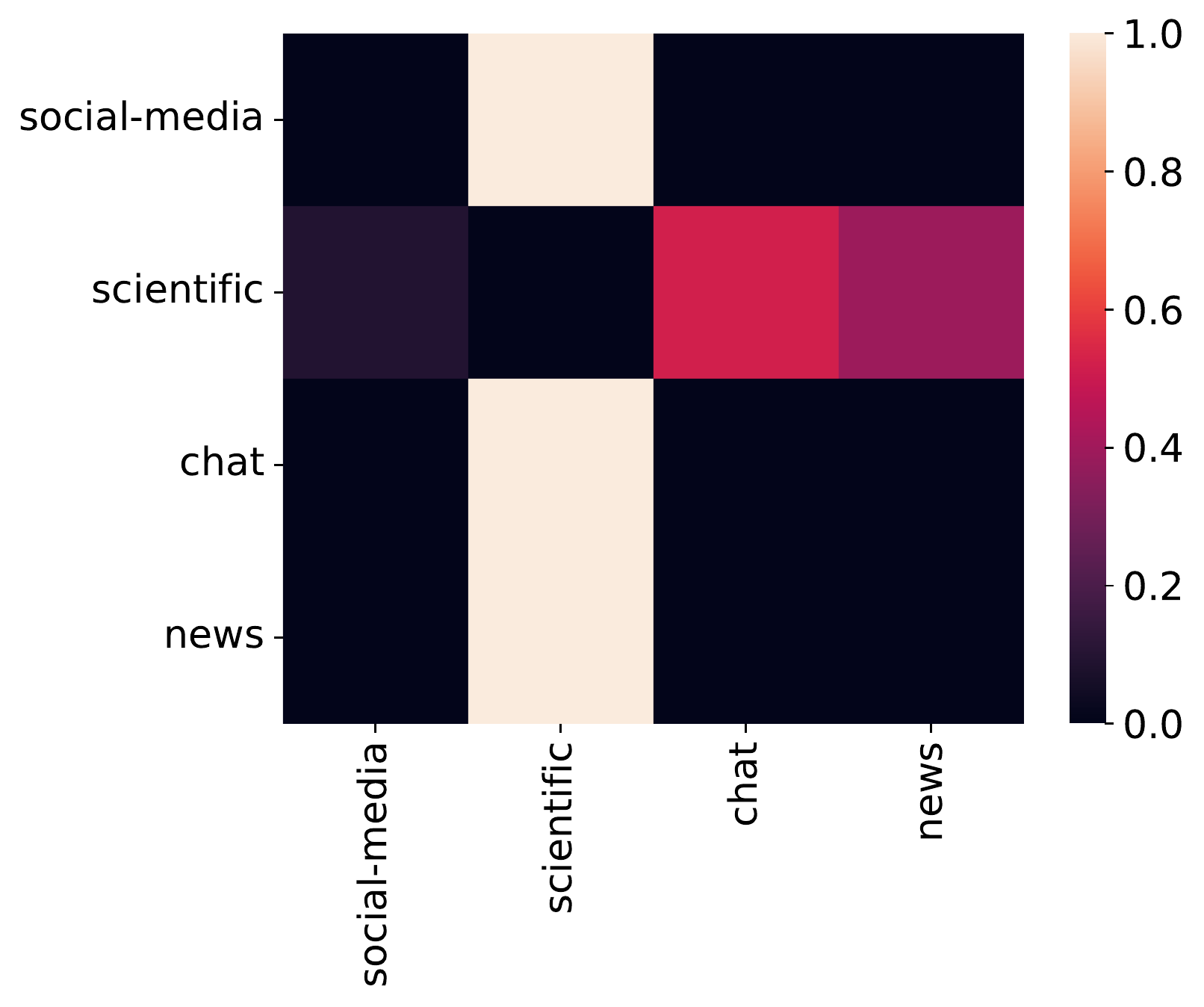}
         \caption{Weights obtained by DAPA for averaging source prefixes.}
         \label{fig: Fig4a}
     \end{subfigure}
     \hspace{5mm}
     \begin{subfigure}[t]{.3\linewidth}
         \includegraphics[width=\textwidth]{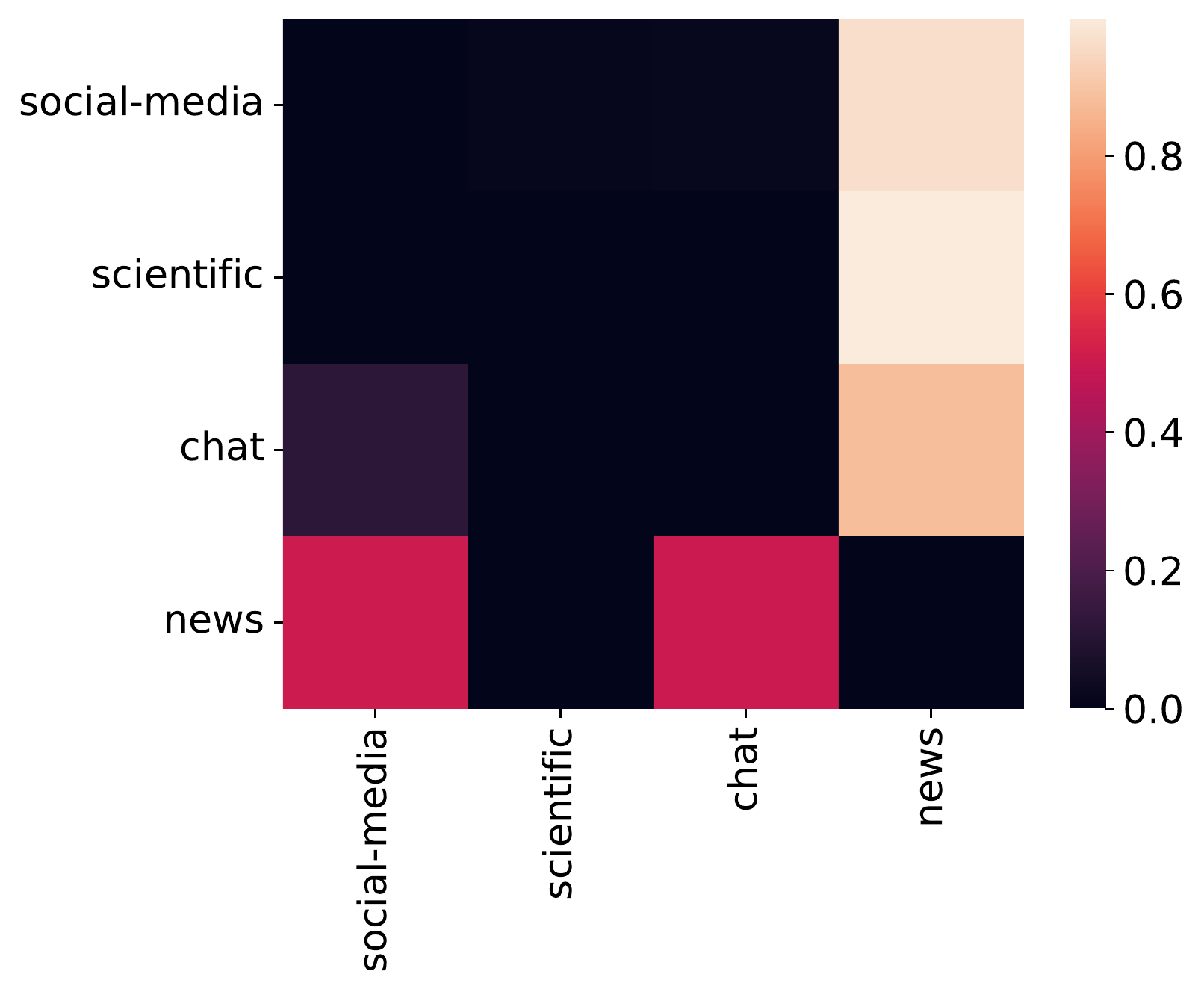}
         \caption{Probabilities assigned to the target domain by a BERT model trained for source domain identification.}
         \label{fig: Fig4b}
     \end{subfigure}
     \hspace{5mm}
        \begin{subfigure}[t]{.3\linewidth}
         \includegraphics[width=\textwidth]{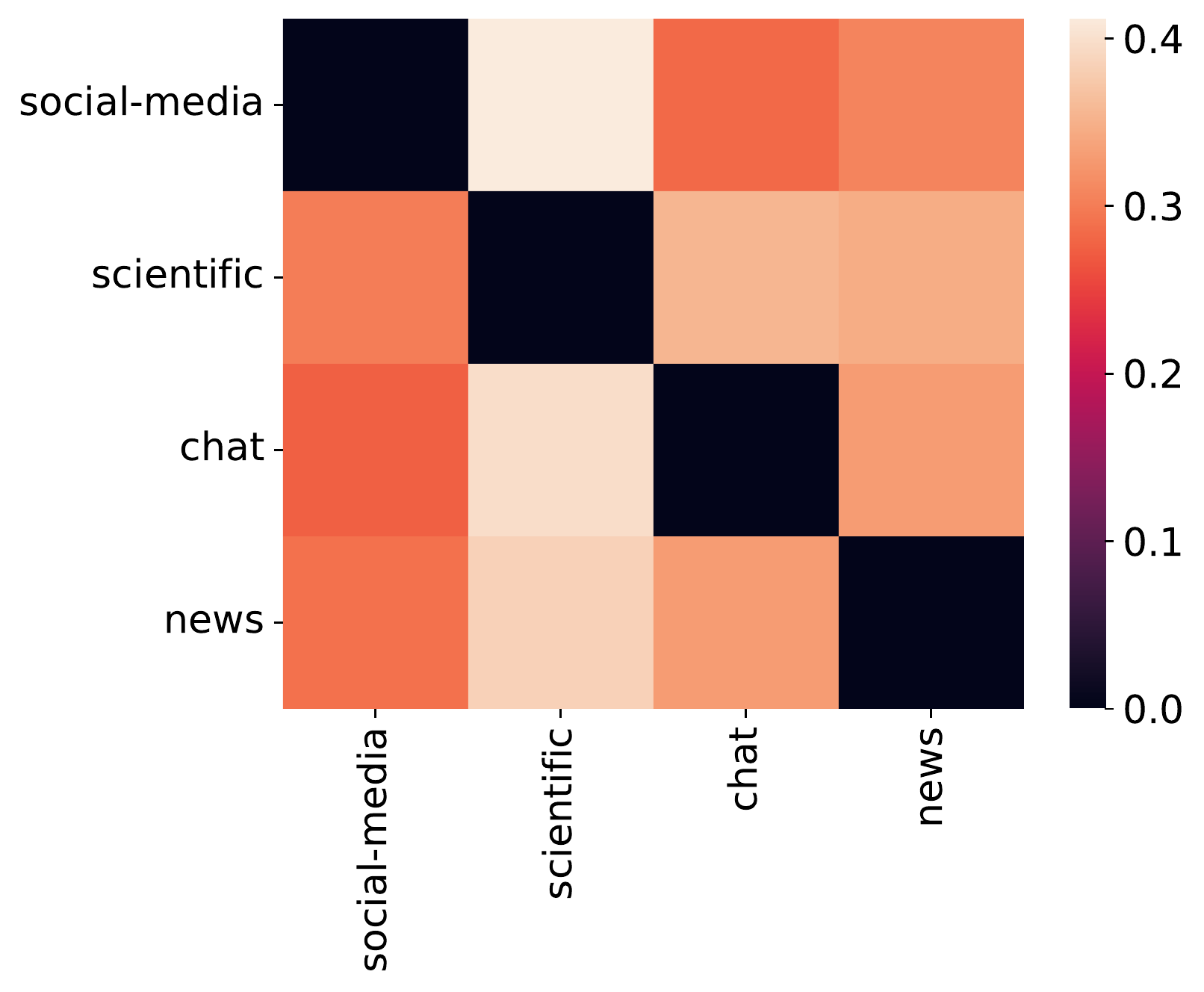}
         \caption{Probabilities assigned to the target domain by DAPA-alt.}
         \label{fig: Fig4c}
     \end{subfigure}
        \caption{Source domain preferences for the target domain obtained through DAPA, a BERT model trained for source domain identification and DAPA-alt.}
        \label{fig:Fig4}
\end{figure*}

On the contrary, ERM-finetune outperforms DAPA on the \textit{scientific} domain.  \citet{Yasunaga2019ScisummNetAL} demonstrate the superior performance of extractive approaches over abstractive summarization approaches for the \textit{scientific} domain. Thus, the ability to copy phrases from the input document becomes imperative to a good performance. Possibly, since all model parameters are tuned for ERM-finetune, its ability to copy phrases from the input document exceeds that of DAPA which only tunes the source prefixes. Despite this, DAPA outperforms the other compared methods for reasons similar to the ones stated previously.

Also, all compared methods outperform DAPA on the \textit{social-media} domain. We find that DAPA allocates a significant amount of weight to the \textit{scientific} domain prefix. However, we observe that the \textit{scientific} domain adversely affects performance on the \textit{social-media} domain (Refer to Section \ref{effect of source domains} for more details). This irregularity may have resulted from the encoder $f$. Further investigation to this is left for future work.  Only $27.94\%$ of the maximal weights selected by DAPA-max belong to the \textit{scientific} domain prefix. Also, DAPA-average assigns equal weights to all the three source domains which is less than the weight assigned by DAPA to the \textit{scientific} domain. Thus both DAPA-average and DAPA-max outperform DAPA. The noise added by DAPA-inst to the weight calculation process results in a smaller weight assigned to the \textit{scientific} domain prefix (0.35 vs 1.00) as a result of which DAPA-inst outperforms DAPA. Owing to the larger dataset size for the \textit{chat} and \textit{news} domains, ERM-prefix and ERM-finetune are less impacted by adverse effects of the \textit{scientific} domain and thus outperform DAPA. ERM-finetune underperforms ERM-prefix probably because of its larger capacity to retain \textit{scientific} domain knowledge.

\section{Analysis} \label{analysis}
To study the impact of various factors in DAPA, and better understand its efficacy, we conduct a series of analysis.

\subsection{Effect of Source Domains on the Target Domain} \label{effect of source domains}
In Table \ref{tab:Table 3}, and Table \ref{tab:Table 4}, we analyze the effect of various source domains on the target domain. Throughout these experiments, $m = 50$ and $C = 50$. Note that the experiments in this section only require recomputation of the prefix averaging weights and thus support the claim that DAPA allows for computationally efficient addition of new source domains. For the \textit{scientific} domain, we can see that the performance is best when only using the \textit{news} domain, in fact, it outperforms ERM-finetune from Table \ref{tab:Table 2}. Adding the \textit{chat} and \textit{social-media} domains only hampers performance. The performance is worst when using only the \textit{social-media} domain. The improvement over this result indicates that DAPA is able to assign appropriate weights to the three source domains allowing for a greater contribution from the \textit{news} and \textit{chat} domains. In real-world applications where the number of domains is significantly greater than our setting, and there is no labelled data to measure the performance over the target domain (to decide the optimal set source prefixes to be averaged), DAPA offers an effective scheme to choose the most appropriate source prefixes.

In the case of the \textit{chat} domain, ablating the \textit{scientific} domain results in degraded performance. On the contrary, excluding the other two prefixes does not affect DAPA's performance. Similarly for the \textit{news} domain, removing the \textit{scientific} domain's prefix results in a significant drop in all three ROUGE scores and removing the \textit{chat} domain results in a slight drop in ROUGE-1 score. Here, we see that both the \textit{chat}, and the \textit{scientific} domain contribute to the performance since using only one of them underperforms the result in Table \ref{tab:Table 2}.

In the case of the \textit{social-media} domain, excluding the \textit{scientific} domain significantly improves DAPA's performance. Also, both the \textit{news} and the \textit{chat} domain contribute to the performance since using only one of them underperforms their weighted average.

\subsection{Effect of Prefix Length $C$}
An analysis on the effect of $C$ is presented in Figure \ref{fig:Fig1}. $C$ most frequent words in the target domain are extracted from $m = 50$ documents. In general, we observe the target domain performance increases up to $C = 50$ following which it either drops or remains more or less the same. Thus, we select $C = 50$ for DAPA. $scientific$ domain ROUGE scores drop significantly after $50$ prefix tokens. We leave further exploration into this for future work.

\begin{table*}[]
\centering
\resizebox{\textwidth}{!}{%
\begin{tabular}{|c|ccc|ccc|ccc|ccc|}
\hline
\multirow{2}{*}{Approach} & \multicolumn{3}{c|}{News} & \multicolumn{3}{c|}{Scientific} & \multicolumn{3}{c|}{Chat} & \multicolumn{3}{c|}{Social-media} \\ \cline{2-13} 

                          & R-1     & R-2     & R-L     & R-1       & R-2       & R-L       & R-1      & R-2    & R-L     & R-1        & R-2       & R-L        \\\hline
DAPA-embed             & 39.06  &  17.96 &   31.81  & 28.92  &  16.79  &  26.24  & 26.32  & 8.21 & 21.31  & 13.04  & \textbf{2.76} & 8.80   \\
DAPA                    & \textbf{40.28}  & \textbf{18.12}  & \textbf{32.78}  & \textbf{30.84}    & \textbf{18.97}    & \textbf{27.23}    & \textbf{28.23}   & \textbf{8.70}  & \textbf{22.86} & \textbf{14.48 }    & 2.68     & \textbf{9.70 }  \\
\hline
\end{tabular}
}
\caption{ROUGE scores for initializing $E^T$ as a weighted average of $E^j$s, i.e. DAPA-embed.}
\label{tab:Table 5}
\end{table*}

\begin{table*}[]
\centering
\resizebox{\textwidth}{!}{%
\begin{tabular}{|c|ccc|ccc|ccc|ccc|}
\hline
\multirow{2}{*}{Approach} & \multicolumn{3}{c|}{News} & \multicolumn{3}{c|}{Scientific} & \multicolumn{3}{c|}{Chat} & \multicolumn{3}{c|}{Social-media} \\ \cline{2-13} 

                          & R-1     & R-2     & R-L     & R-1       & R-2       & R-L       & R-1      & R-2    & R-L     & R-1        & R-2       & R-L        \\\hline
DAPA-alt              & 36.12  &  15.96 &  30.70  & 29.42  &  18.02  & 27.32 & 23.57 & 5.84  & 20.05 & \textbf{16.81}  & \textbf{2.90}  &  \textbf{11.82}  \\
DAPA                     & \textbf{40.28}  & \textbf{18.12}  & \textbf{32.78}  & \textbf{30.84}    & \textbf{18.97}    & \textbf{27.23}    & \textbf{28.23}   & \textbf{8.70}   & \textbf{22.86 } & 14.48     & 2.68     & 9.70   \\
\hline
\end{tabular}
}
\caption{ROUGE scores for an alternative way of computing $w^j$s as discussed in Section \ref{Alternative}, i.e. DAPA-alt.}
\label{tab:Table 6}
\end{table*}

\subsection{Effect of $m$ on $W$}
An analysis on the effect of the number of sentences used for computing weights to average source prefixes is presented in Figure \ref{fig:Fig 2}. Beyond $m = 20$, the performance on the target domain remains more or less constant. Thus, we stick to our initial choice of using $50$ sentences to compute weights $W$.

An analysis on the effect of the number of sentences used for obtaining $C$ most frequent tokens is presented in Figure \ref{fig:Fig3}. Again, the performance does not vary significantly beyond $m = 20$. Thus, we hold to our initial choice of $m = 50$ for our main experiments. Note that we do not include results on the \textit{scientific domain} for this subsection since its test set has only $10$ instances, and we use all of them for our experiments. 

\subsection{Does Averaging over Source $E^j$s Help?}
In our main method, we initialize $E^T$ with $C$ most frequent sentencepiece tokens from the $m$ unlabelled documents. Here, we explore an alternative way of initializing $E^T$ wherein we use $w^j$s to take a weighted average of source $E^j$s :
\begin{equation}
    E^{T} = \sum_{j = 1}^{n} w^j E^j
\end{equation}

Results for this initialization scheme (DAPA-embed) are presented in Table \ref{tab:Table 5}. DAPA outperforms DAPA-embed across domains demonstrating the benefits of supplying $P^T$ with some prior target domain knowledge by initializing $E^T$ with $C$ most frequent sentencepiece tokens from the $m$ unlabelled documents.

\subsection{Does Softmax before Summation Help?}
\label{Alternative}
Here, we propose an alternative way of computing $w^j$. Instead of summing over the document-summary cosine similarities (Equation \ref{eq 4}) and then applying the softmax operation (Equation \ref{eq 5}), we first apply the softmax operation to document-summary similarity scores following which we average over them. That is, we replace Equation \ref{eq 4} with:
\begin{equation}
       w_{i}^j = \frac{\text{exp}(\text{cosine-similarity}(r_{i}^{j}, t_{i}))}{\sum_{k = 1}^{n} \text{exp}(\text{cosine-similarity}(r_{i}^{k}, t_{i}))}
\end{equation}
and Equation \ref{eq 5} with
\begin{equation}
       w^j = \frac{1}{m} \sum_{i=1}^{m} w_{i}^j
\end{equation}
By doing so, we are flattening the weights $w^j$. This is evident from Figure \ref{fig: Fig4c} where the target domain assigns near equal weights to all three source domains. This is different from DAPA, where the weights $w^j$ are sharp as depicted in Figure \ref{fig: Fig4a}. In Table \ref{tab:Table 6}, DAPA outperforms this alternative strategy (DAPA-alt) on three out of the four domains. A flattened $w^j$ distribution approaches DAPA-average, and thus, does not benefit from DAPA's weight averaging scheme. DAPA-alt outperforms DAPA on the \textit{social-media} domain since it assigns near equal weights to each source domain (Refer to Section \ref{main results} for a detailed analysis).

\subsection{Is DAPA correlated to Document Similarity?}
We analyze how DAPA's weight assigning process aligns with source-target domain similarity at the document level. For this we train BERT-base \citep{DBLP:conf/naacl/DevlinCLT19} on 300 source domain documents (100 from each source domain) for source domain identification. We evaluate this model on  the target domain's $D^{T}_{50, sample}$. We plot the average probabilities assigned to each source domain in Figure \ref{fig: Fig4b}. Also, in Figure \ref{fig: Fig4a}, we plot weights $W$ computed by DAPA. Note that BERT-base achieves perfect test accuracy when evaluated on an in-domain validation set for each training setting. Entries of the form $[x, x]$ are always zero owing to our domain generalization setting.

From the two plots, it is clear that DAPA's weight assignment process does not always correlate with source-target domain similarity at the document level. For the \textit{news} domain, as per Table \ref{tab:Table 3} and Table \ref{tab:Table 4}, the \textit{scientific} domain contributes most to the model's performance, on the other hand, BERT assigns near equal probability to the \textit{social-media} and \textit{chat} domain, and assigns near zero probability to the \textit{scientific} domain. Similarly, for the \textit{chat} domain, the \textit{scientific} domain is vital to good performance, however, BERT assigns a low probability to it. Whereas, for the \textit{social-media} and \textit{scientific} domain, BERT does a better job and assigns a higher probability to the \textit{news} domain. These results navigate us to the conclusion that DAPA does not use document level similarities and indeed relies on the summary generation capabilities of source prefixes.

\section{Conclusion} \label{Conclusion}
In this paper, we present DAPA, a lightweight, domain aligned prefix averaging approach to domain generalization in abstractive summarization. DAPA utilizes source prefixes to generate summaries for a small number of target domain documents. The similarity of these summaries to their corresponding documents are used for calculating weights required to average source prefixes. DAPA can easily account for the addition of new source domains since only the prefix averaging weights need to be recomputed. On four diverse summarization domains, DAPA either performs comparably or outperforms the baselines. We also perform an in-depth analysis of various components of DAPA to further strengthen our design choices. In future, we would like to develop an improved similarity function $f$ and analyze the loss landscapes of these models to corroborate our prefix averaging strategy.

\section{Limitations} \label{Limitations}
Our work focuses on domain generalization for abstractive summarization through prefix averaging. However, we do not experiment with larger backbone models due to computational constraints. Based on previous works we expect our approach's performance to improve with model size. Also, a larger sequence length for prefix tuning increases the computational costs at inference. 

Another limitation of our work is that we do not test it on natural language understanding tasks. This can be part of a future work.

\section{Ethical Statement} \label{Ethics}
We consider our approach to have low ethical risks since we do not utilize any data biases. Our approach could be extended to any natural language generation task and does not constraint the input/output structure. We therefore conclude that our method would not bring any harmful ethical impact.

\bibliography{custom}
\bibliographystyle{acl_natbib}

\appendix
\section{Additional Results}
\label{Additional Results}
\begin{table*}[]
\centering
\resizebox{\textwidth}{!}{%
\begin{tabular}{|c|ccc|ccc|ccc|ccc|}
\hline
\multirow{2}{*}{Approach} & \multicolumn{3}{c|}{News} & \multicolumn{3}{c|}{Scientific} & \multicolumn{3}{c|}{Chat} & \multicolumn{3}{c|}{Social-media} \\ \cline{2-13} 

                          & R-1     & R-2     & R-L     & R-1       & R-2       & R-L       & R-1      & R-2    & R-L     & R-1        & R-2       & R-L        \\\hline
Finetune-target &  37.65 & 17.05 & 32.06 & 39.66 & 21.76 & 35.57 & 32.65 & 12.43 & 29.03 & 17.38 & 2.92 & 12.77 \\
Prefix-target & 38.10 &  17.61 & 33.11 & 49.88 & 30.64 & 43.19 & 39.30 & 15.67 & 33.49 & 19.48 & 4.10 & 15.35 \\
Finetune & 41.20 & 19.18 & 35.20 & 52.97 & 32.27 & 45.85 & 49.42 & 24.31 & 42.43 & 25.06 & 7.23 & 19.96  \\
Prefix & 40.56 & 19.07 & 34.94 & 49.58 & 29.72 & 42.84 & 46.20 & 22.48 & 40.27 & 23.22 & 6.30 & 18.59 \\
DAPA                     & 40.28  & 18.12  & 32.78  & 30.84    & 18.97    & 27.23    & 28.23   & 8.70   & 22.86 & 14.48     & 2.68     & 9.70   \\
\hline
\end{tabular}
}
\caption{Results for the four additional baselines.}
\label{tab:Table end}
\end{table*}

For the sake of completeness, we present four additional baselines in Table \ref{tab:Table end}. Finetune-target assumes that the $m$ target domain documents are labelled and thus uses standard finetuning to update T5-small parameters with these $m$ documents. Likewise, Prefix-target uses prefix tuning to train a prefix from scratch with these $m$ documents. Since, the above two baselines use labelled documents they cannot be directly compared with DAPA and only present an upper bound to the performance of DAPA. Using $m$ labelled documents reaps benefits on three out of the four domains. Note that we use $m = 50$ for the above two baselines.

We also present results for full finetuning (Finetune) and full prefix tuning (Prefix) on each of the four domains. Finetune uses the target domain's entire training set to update T5-small parameters and Prefix trains a prefix from scratch on the target domain's training set. Again, these two baselines only act us upper bounds to DAPA and outperform it across domains.

\end{document}